\documentclass[11pt]{article}
\usepackage[english]{babel}

\usepackage[letterpaper,top=2cm,bottom=2cm,left=3cm,right=3cm,marginparwidth=1.75cm]{geometry}

\usepackage{amsmath}
\usepackage{graphicx}
\usepackage[colorlinks=true, allcolors=blue]{hyperref}

\usepackage{authblk}

\title{Privacy Amplification via Random Participation\\ in Federated Learning}
\author{Burak Hasırcıoğlu}
\author{Deniz Gündüz}
\affil{Information Processing and Communications Lab, Imperial College London, UK}
\affil{E-mail: \{b.hasircioglu18, d.gunduz\}@imperial.ac.uk}

\date{}

\usepackage{microtype}
\usepackage{graphicx}
\usepackage{subfigure}
\usepackage{booktabs} 

\usepackage{hyperref}

\usepackage{amsmath}
\usepackage{amssymb}
\usepackage{mathtools}
\usepackage{amsthm}

\usepackage[capitalize,noabbrev]{cleveref}

\theoremstyle{plain}
\newtheorem{theorem}{Theorem}[section]

\newtheorem{corollary}[theorem]{Corollary}
\theoremstyle{definition}
\newtheorem{definition}[theorem]{Definition}

\theoremstyle{remark}

\usepackage{xspace}
\usepackage{comment}
\newcommand{\OnlyLocal}{Only Local Sampling\xspace}
\providecommand{\OnlyLocalAcr}{OLS\xspace}
\providecommand{\UpperBound}{Weak Client Sampling\xspace}
\providecommand{\UpperBoundAcr}{WCS\xspace}
\providecommand{\LowerBound}{Centralized Shuffling\xspace}
\providecommand{\LowerBoundAcr}{CS\xspace}

\usepackage{tikz}
\usepackage{pgfplots}
\usepackage{pgfgantt}

\usepackage{algorithm}
\usepackage{algorithmic}
\usepackage[normalem]{ulem}

\begin{document}
\maketitle

\begin{abstract}
Running a randomized algorithm on a subsampled dataset instead of the entire dataset amplifies differential privacy guarantees. In this work, in a federated setting, we consider random participation of the clients in addition to subsampling their local datasets. Since such random participation of the clients creates correlation among the samples of the same client in their subsampling, we analyze the corresponding privacy amplification via non-uniform subsampling. We show that when the size of the local datasets is small, the privacy guarantees via random participation is close to those of the centralized setting, in which the entire dataset is located in a single host and subsampled. On the other hand, when the local datasets are large, observing the output of the algorithm may disclose the identities of the sampled clients with high confidence. Our analysis reveals that, even in this case, privacy guarantees via random participation outperform those via only local subsampling.
\end{abstract}

\section{Introduction}
Federated learning (FL) framework allows several clients to collaboratively and iteratively learn from each other's data with the coordination of a parameter server (PS) \cite{mcmahan17a}. In a typical scenario, the PS averages the model updates received from clients based on their local data. Then, it updates the global model and broadcasts the updated global model back to clients for the next iteration. In this learning model, since the data itself never leaves the participating clients, in terms of privacy, the FL framework is usually perceived as superior to centralized training, in which entire data is offloaded to a central server. Although this is true to some extent, many recent works \cite{melis2019exploiting,zhu2019deep,geiping2020neurips,carlini2019secret,salem2020updates} have shown that the model updates from the clients as well as the final deployed model can leak many features of the clients' local datasets, including reconstruction of some data samples used for training. Therefore, additional mechanisms with formal and quantifiable privacy guarantees need to be employed in FL.

Differential privacy (DP) \cite{dwork2014algorithmic} is the gold standard quantifying the privacy leakage in privacy-preserving data analysis tasks. It is a measure of the indistinguishability between two outputs of an algorithm when two neighboring datasets are fed into the algorithm as inputs. A deterministic algorithm can be made differentially private by randomizing its output as long as the introduced randomness is independent from the input and secret to the adversary. Output perturbation via an additive noise such as Gaussian or Laplace noise \cite{dwork2006calibrating} is a common example of such randomization techniques. However, there exist other sources of randomness that do not guarantee indistinguishability alone, but they can amplify the final DP guarantees when they are cascaded with techniques already guaranteeing DP. \emph{Subsampling} \cite{chaudhuri2006random, balle2018privacy, wang2019subsampled, mironov2019r} and \emph{shuffling} \cite{erlingsson2019amplification, feldman2020hiding} are among the most prominent of such tools. In subsampling, a random subset from the original dataset is sampled to be used as the input to the privacy-preserving mechanism. In shuffling, on the other hand, the outputs resulting from different inputs are randomly shuffled so that mapping between the inputs and the outputs are masked, resulting in anonymized outputs. Such privacy amplification techniques further confuse an adversary when used in conjunction with output perturbation techniques, and the amplified privacy guarantees are achieved without sacrificing the utility.

In this work, we consider a FL setting with distributed stochastic gradient descent (DSGD) using a trusted PS, in which formal DP guarantees are achieved for all intermediate models including the final deployed one. For this, we update the global model by the average of the gradients collected from participating clients, which is perturbed via an additive Gaussian noise, following the seminal work of Abadi et al. \cite{abadi2016deep}. To further amplify the privacy guarantee, we consider two types of sampling: (1) client sampling, and (2) local dataset sampling. In each iteration, first, available clients during that iteration randomly decide whether to participate or not, independently from the decisions of other clients. Then, each client that decides to participate samples a subset from its local dataset such that each element is sampled independently from the other elements in the local dataset. That is, in both sampling phases, we employ Poisson sampling separately. Analysing the privacy amplification guarantee of this sampling technique in a federated setting is  challenging. More specifically, random participation of the clients introduces correlation between the elements located in a single client in their sampling. For example, given an element is sampled from a specific client, the probability of another element being sampled from the same client is larger than the probability of another element from another client being sampled. Therefore, employing random participation of clients and local dataset sampling together poses a non-uniform sampling problem, and in this work, we analyse the central DP guarantees of such a setting.

\subsection{Related Work and Motivation}
\label{sec:motivation}

For FL scenarios with many participants, sampling the participating clients is essential for efficient use of the power and the communication resources even when privacy is not a constraint. Consider a learning problem with millions of mobile devices. Thousands of iterations are carried out in a typical learning task, and if each mobile device participates in every iteration, the power spent on the computations and the extra communication to transmit the results to the PS will result in a very large overall energy cost. In addition, communication overhead of such a dense participation may cause congestion in the communication networks; and thus, is not scalable. Moreover, for the learning tasks employing mini-batch optimization techniques, limited number of samples per iteration is preferable. Hence, client sampling can help to reach the targeted batch size in addition to the local dataset sampling. All these aforementioned benefits make client sampling one of the indispensable components of FL, whose effect on the convergence and final performance of FL algorithms have been widely studied \cite{chen2020optimal,ruan2021towards,cho2020client}. In addition to these benefits, since it brings additional randomness to the training procedure, it is expected to bring some additional amplification to the DP guarantees. Hence, rather than proposing the inclusion of a new mechanism to amplify DP guarantees, in this work, our goal is to utilize an already-employed mechanism to improve the privacy guarantees of the task.

Differentially private deep learning in a centralized setting is studied in \cite{abadi2016deep}. The authors employ Poisson subsampling of the dataset and the gradient of every sample is clipped and Gaussian noise is added to it. They further introduce the moments accountant technique to calculate the composition of the total privacy leakage throughout the iterations. Similar techniques to those in \cite{abadi2016deep} are extended to the federated setting in \cite{geyer2017differentially, mcmahan2018learning, seif2021privacy}. In these works, similarly to our approach, client sampling is considered, but the achieved DP guarantees are at the client-level, which means the participation of a client is indistinguishable from its non-participation. Although client-level DP guarantees are meaningful when the local dataset is composed of the elements from the same source, e.g., personal photos stored in a mobile phone, in some cases, it might be too conservative. To achieve DP guarantees protecting all the elements in a client, a larger amount of noise must be introduced compared to sample-level guarantees, and hence, client-level guarantees result in a larger loss in utility. Moreover, in some cases, the datasets stored by the clients may comprise of samples from different individuals, hence, it may not be necessary to protect client-level privacy. For instance, while learning from medical data, each client may represent a hospital and sample-level DP guarantees would suffice \cite{malekzadeh2021dopamine}.

Sample-level DP in the federated setting with client sampling is studied in \cite{balle2020privacy,girgis2021shuffled,girgis2021differentially} together with shuffling. In the analysis carried out in \cite{balle2020privacy}, each client is assumed to store only one sample, in which case the client-level guarantees are equivalent to sample-level guarantees. While each client is allowed to store more than one sample in \cite{girgis2021shuffled,girgis2021differentially}, the number of sampled elements is determined in advance. In \cite{girgis2021differentially}, only one element is sampled from the local dataset at each iteration, while more than one but a constant number of elements are sampled in \cite{girgis2021shuffled}. It is also worth noting that different from \cite{balle2020privacy} and the current paper, in \cite{girgis2021shuffled,girgis2021differentially}, communication efficiency is studied together with privacy, and a random compression mechanism is used as the randomizer. In all of these works \cite{balle2020privacy,girgis2021shuffled,girgis2021differentially}, the clients are assumed to employ a local randomizer that satisfies pure DP, which we will formally define in \cref{sec:preliminaries}, and the privacy analysis is based on shuffling the local responses. Such an assumption of pure DP for the local randomizers turns out to be useful while jointly analysing client sampling and shuffling. However, if the local randomizers provide approximate DP guarantees instead of pure DP, which is the case when Gaussian noise is used as the randomizer, central privacy guarantees may be substantially degraded. Therefore, novel methods to jointly analyze privacy amplification via client sampling and local dataset sampling that do not rely on shuffling are needed, and this constitutes the main motivation and contribution of our work.

\subsection{Our Contributions} The major contributions of our work and its novelty with respect to the current literature can be summarized as follows:

\begin{itemize}
\item We propose a client sampling algorithm for privacy amplification in FL. Each available client randomly decides to participate or not at each iteration. Having decided to participate, each client then samples its local dataset randomly. 
\item Since each client decides to participate or not independently from any other external factor, it can account for its own privacy loss after each iteration without requiring any other iteration-specific information about the system, such as the number of available clients, the number of sampled clients, or the number of sampled elements by the participating clients. This local sampling and privacy accounting mechanism makes the implementation of our algorithm feasible in practice.
\item We provide a theoretical analysis of the central privacy guarantees of the proposed algorithm. Since random client participation leads to a non-uniform sampling among the data points stored at different clients, the analysis is non-trivial and cannot be directly obtained from the standard privacy amplification results via subsampling. Unlike the literature, our analysis does not rely on shuffling, and we do not need local randomizers with pure DP guarantees. This is especially useful when Gaussian noise is used as a randomizer.
\end{itemize}

\section{Preliminaries}\label{sec:preliminaries}

In this section, we review some preliminary notions
about DP and subsampling, which will be used in the discussion of our algorithm and its privacy analysis.

\begin{definition}[Differential Privacy]
A randomized mechanism $\mathcal{M}:\mathcal{D}\to\mathcal{S}$ is
$(\varepsilon,\delta)$ differentially private (DP) if 
\[
\Pr[\mathcal{M}(D)\in S]\leq e^{\varepsilon}\Pr[\mathcal{M}(D')\in S]+\delta,
\]
for all neighboring datasets $D$ and $D'$, i.e., sets differing in only
one element, and $\forall S\subset \mathcal{S}$, where $\varepsilon>0$ and $\delta\in[0,1)$. If $\delta =0$, then $\mathcal{M}$ is called $\varepsilon$ pure differentially private, or $\varepsilon$-DP.
\end{definition}

The neighboring relation between $D$ and $D'$ depends on the context.
In our work, we assume $D$ can be generated from $D'$ by either removing or adding
one element. $\delta$ in the definition above quantifies
the failure probability of the relation $\Pr[\mathcal{M}(D)\in\mathcal{S}]\leq e^{\varepsilon}\Pr[\mathcal{M}(D')\in\mathcal{S}]$. Thus, $(\varepsilon,\delta)$-DP
is, in fact, a relaxation of $\varepsilon$-DP. 

Alternative to the
above definition, $(\varepsilon,\delta)$-DP can be equivalently expressed
in terms of \emph{hockey stick divergence}, which we introduce
next.

\begin{definition}[\cite{sason2016f}]\label{def:hockey-stick}
We define the \emph{hockey stick divergence }between two probability
measures $\mu$ and $\mu'$ as 
\begin{equation}
D_{\alpha}(\mu||\mu')\triangleq\int_{Z}\left[d\mu(z)-\alpha d\mu'(z)\right]_{+}d(z)\label{eq:hockey-stick}
\end{equation}
where $\left[\cdot\right]_{+}\triangleq\max\{0,\cdot\}$.
\end{definition}

Based on hockey stick divergence, $(\varepsilon,\delta)$-DP can be expressed as follows.
\begin{theorem}[\cite{barthe2013beyond}, Theorem 1 in \cite{balle2018privacy}]
\label{thm:privacy_profile}A mechanism $\mathcal{M}$ is $(\varepsilon,\delta)$-DP
if and only if $\sup_{D,D'}D_{\alpha}(\mathcal{M}(D)||\mathcal{M}(D'))\leq\delta$,
where $D$ and $D'$ are two neighboring datasets and $\alpha=e^{\varepsilon}$.
\end{theorem}
From \cref{thm:privacy_profile}, we can conclude that there is a trade-off
between $\varepsilon$ and $\delta$ and we can perceive $\delta$
as a function of $\varepsilon$.

Gaussian mechanism is commonly used to achieve differential privacy guarantees for machine learning tasks \cite{abadi2016deep}. The following theorem formally characterizes its guarantees.

\begin{theorem}[Theorem 8 in \cite{balle2018improving}]\label{thm:gaussian_mechanism}
Let $f:\mathcal{D}\to \mathbb{R}^m$ be a function with $||f(D)-f(D')||_2\leq C$, where $D$ and $D'$ are neighboring datasets and $||\cdot||_2$ denotes the $L_2$ norm. A mechanism $\mathcal{M}(D)=f(D)+\mathcal{N}(0,\sigma^2)$ is $(\varepsilon,\delta)$-DP if and only if 
\begin{equation}
    \Phi\left(\frac{C}{2\sigma}-\frac{\varepsilon\sigma}{C}\right)-e^{\varepsilon}\Phi\left(-\frac{C}{2\sigma}-\frac{\varepsilon\sigma}{C}\right)\leq \delta,
\end{equation}
where $\Phi$ is the cumulative distribution function (CDF) of the standard normal distribution.
\end{theorem}

Next, we present another result from \cite{balle2018privacy}, which characterizes
the privacy amplification via subsampling. 
\begin{theorem}[Theorem 8 in \cite{balle2018privacy}]
\label{thm:only-local-sampling}Let $\mathcal{M}$ be a randomized
mechanism satisfying $(\varepsilon,\delta_{\mathcal{M}}(\varepsilon))$-DP.
We define $\mathcal{M}'$ as the randomized mechanism when the input
is first sampled from a larger dataset with Poisson sampling with probability
$q$, then $\mathcal{M}$ is applied on the sampled subset. Then, we have 
$\delta_{\mathcal{M}'}(\varepsilon')\leq q\delta_{\mathcal{M}}(\varepsilon)$,
where $\varepsilon'=\log\left(1+q(e^{\varepsilon}-1)\right)$.
\end{theorem}
Note that this result is only valid for uniform sampling, i.e., the cases in which a subset is sampled from a larger dataset such that each element is sampled independently and identically distributed (i.i.d.).
with probability $q$. Hence, it does not consider any correlation
between the sampled elements. However, it serves as a baseline
to evaluate the performance of our scheme.

\section{Problem Setting}

In our setting, we consider a FL scenario with $N$ clients and a PS. Client $i\in[N]$\footnote{We define $[N]\triangleq\{1,2,\dots,N\}$.}
has a dataset $D_{i}$ such that $D=\cup_{i\in[N]}D_{i}$.
For simplicity, we assume that each client holds the same number of data samples, i.e., $|D_{i}|=d$, $\forall i\in[N]$, but our analysis holds for different $|D_i|$ values as well. We allow one of the clients to violate this assumption to satisfy the neighboring
relation, i.e., at most one of the clients can store a dataset of size $d+1$. We consider differentially private distributed stochastic
gradient descent (DP-DSGD) in a FL setting. Namely,
the clients and the PS collectively learn a model $\boldsymbol{w}\in \mathbb{R}^m$
through $T\in \mathbb{Z}^+$ iterations by minimizing the loss function 
\begin{equation}
    \hat{\ell}(\boldsymbol{w},D)=\frac{1}{N}\sum_{i\in[N]}\frac{1}{|D_i|}\sum_{s\in D_i}\ell(\boldsymbol{w},s),
\end{equation}
where $\ell$ is a general loss function depending on
the nature of the learning problem. In general, we allow it to be non-convex; hence, our results are applicable to deep neural networks. Since clients can go offline for several reasons in FL, in each iteration $t$, we assume $N_t\leq N$ clients are available and $N_t$ is known to the PS but not necessarily to the clients.

\subsection{Threat Model}

Following \cite{balle2020privacy,girgis2021shuffled, girgis2021differentially}, we assume a trusted PS and honest but curious clients, i.e., they
do not deviate from the protocol but they can try to learn about the
local datasets of other clients. We further assume the presence of secure communication channel between each client and the PS such that neither the presence
of communication nor the content of the
transmitted messages is disclosed to third parties. Therefore, in the case of random participation, for an adversary, it is not possible to infer the participating clients by tapping into the channels.

Trusted PS assumption may sound strong, but due to the nature of the client sampling problem, it is essential. Keeping the identities of the participating clients secret from the PS is difficult without employing a trusted third party. Hence, as done in similar works \cite{balle2020privacy,girgis2021shuffled,girgis2021differentially}, we make this assumption. This assumption is valid for the scenarios in which the training is done with the help of a trusted PS, and the model is publicly deployed after training. If the PS is not trusted, client sampling can still be employed for a price of increased communication cost, and our analysis still holds. We elaborate on this in \cref{subsec:untrusted_ps}.

\section{Main Results}\label{sec:main_results}
In this section, we first present the details of the proposed algorithm and then we state its privacy guarantees.

\subsection{Algorithm Description}
At the beginning of each iteration $t\in[T]$, independently from other clients, each client
randomly decides whether to participate or not in that
iteration, with probability $p$. If a client decides to participate,
then it further samples a subset from its local dataset such that each element is sampled i.i.d. with
probability $q$. Let the set of sampled clients in iteration $t$ be $P_t$ and the set of sampled elements in client $i\in P_t$ be $S_{i,t}$. After sampling, client $i\in P_t$ computes the gradients $\nabla\ell(\boldsymbol{w},s)$
for all $s\in S_{i,t}$. If the $L_2$ norm of the gradient of any sample is greater than
a predetermined constant $C$, then it is scaled down to $C$ to guarantee $||\nabla\ell(\boldsymbol{w},s)||_2\leq C$. Then, the client $i$ aggregates
all the sample gradients, and sends
the sum to the PS. The PS further aggregates all the summations from the participating clients and adds a Gaussian noise $\mathcal{N}(0,\sigma^{2}\mathbf{I}_m)$ to the sum, where $\mathbf{I}_m$ is the identity matrix with dimension $m$. Then, it scales the noisy sum by $\frac{1}{pN_tqd}$
to have an unbiased estimate of the gradient. The final expression after these operations at the PS is 
\begin{equation}
    \hat{\boldsymbol{g}}_{t}=\frac{1}{pN_tqd}\left(\sum_{i\in P_t}\sum_{s\in S_{i,t}}\nabla\ell(\boldsymbol{w},s)+\mathcal{N}(0,\sigma^{2}\mathbf{I}_m)\right).
\end{equation}
Finally, the PS updates the model as $\boldsymbol{w}_{t+1}=\boldsymbol{w}_{t}-\eta_t\hat{\boldsymbol{g}}_{t}$, where $\eta_t$ is the learning rate for the iteration $t$. 
The pseudo-code for our procedure is given in \cref{alg:DP-DSGD}.

\begin{algorithm}
\begin{small}
\begin{center}
\begin{algorithmic}
\hspace{-8pt}\textbf{Protocol in client $i$:}
\FOR{$t \in [T]$}
\STATE Sample $B\sim Bern(p)$
\IF{$B=1$} 
\STATE Inform the PS that $B=1$
\STATE Receive $\boldsymbol{w}_t$ from the PS
\STATE Initialize $\boldsymbol{g}_i=0$
\STATE Sample $S_i$ from $D_i$, w.p. $q$, i.i.d. for each sample
\FOR{$s\in S_i$}
\STATE Increment $\boldsymbol{g}_i$ by $\nabla\ell(\boldsymbol{w}_t,s)/\max\left\{1,\frac{||\nabla\ell(\boldsymbol{w}_t,s)||_2}{C}\right\}$
\ENDFOR
\STATE Send $\boldsymbol{g}_i$ to PS
\ENDIF 
\ENDFOR
\end{algorithmic}
\end{center} 

\begin{center}
\begin{algorithmic}
\hspace{-8pt}\textbf{Protocol in the PS:}
\FOR{$t\in[T]$}
\STATE Learn $P_t$ from the clients
\STATE Broadcast $\boldsymbol{w}_t$ to $\forall i\in P_t$
\STATE Receive $\boldsymbol{g}_i$ from $\forall i \in P_t$
\STATE Aggregate, randomize and scale:\\ \hspace{10pt} $\hat{\boldsymbol{g}}_t=\frac{1}{pN_tqd}\left(\sum_{i\in P_i}{g}_i + \mathcal{N}(0,\sigma^2\mathbf{I}_m)\right)$
\STATE Update $\boldsymbol{w}_{t+1}=\boldsymbol{w}_{t}-\eta_t\hat{\boldsymbol{g}}_{t}$
\ENDFOR
\end{algorithmic}
\end{center}
\end{small}
\caption{DP-DSGD with random participation\label{alg:DP-DSGD}}
\end{algorithm}

\subsection{Privacy Guarantees}
In the following theorem, we present the privacy guarantees of DP-DSGD with random participation.

\begin{theorem}\label{thm:main}
Each iteration of DP-DSGD with random participation algorithm is $(\varepsilon,\delta)$-DP,
where, for any $\varepsilon'>0$,
\begin{equation}
\varepsilon=\log\left(1+pq\left(e^{\varepsilon'}-1\right)\right),    
\end{equation}

\begin{multline}
\delta=pq\Bigg(1-\alpha'\bar{c_2}+\alpha'\bar{c_1}\left(\Phi_{0,\sigma}(z^*)-1\right)
+\sum_{i=0}^{d}\binom{d}{i}q^{i}(1-q)^{d-i}\Big(\alpha'\bar{c_2}\Phi_{iC,\sigma}(z^*)-\Phi_{(i+1)C,\sigma}(z^*)\Big)\Bigg),
\end{multline}
$\bar{c_1} \triangleq \left(1-e^{\varepsilon-\varepsilon'}\right)\frac{1-p}{1-pq}$, $\bar{c_2}\triangleq \frac{p(1-q)}{1-pq}\left(1-e^{\varepsilon-\varepsilon'}\right)+e^{\varepsilon-\varepsilon'}$, $\Phi_{C,\sigma}(z^{*})$
is the CDF of the Gaussian distribution with mean $C$ and standard
deviation $\sigma$, and $z^{*}$ is the solution, for $z$, of the following equation: 
\begin{multline}
\sum_{i=0}^{d}\binom{d}{i}q^{i}(1-q)^{d-i}\Big(\mathcal{N}(z,(i+1)C,\sigma^{2}) -\alpha'\bar{c_{2}}\mathcal{N}(z,iC,\sigma^{2})\Big)-\alpha'\bar{c_{1}}\mathcal{N}(z,0,\sigma^{2})=0.\label{eq:z_star_eq}
\end{multline}

\end{theorem}
Note that although the analytical solution of \cref{eq:z_star_eq} is
not tractable, it can be efficiently solved numerically. We provide a more
detailed discussion on this and the proof of \cref{thm:main}  in \cref{sec:proofs}.

\section{Discussion and Numerical Results}\label{sec:discussion_and_numerical}

First, we would like to emphasize that the number of available clients $N_t$ and the sizes of the datasets $d$ at the clients are needed only at the PS to correctly scale the average of the gradients. They are not needed to compute the privacy loss, and hence, each client can keep track of its own privacy loss without needing any iteration-specific information about the system.

In our algorithm, the noise is added by the PS centrally, and the variance of the noise does not depend on any iteration-specific information. Since the number of sampled clients in each iteration can vary, central noise addition makes each client keep track of its own privacy loss much more easier in the FL setting without the need to know the number of sampled clients in each iteration. Central noise addition might be seen as a disadvantage of our scheme since there is no local randomizers at the clients, but in case of a trusted PS, our privacy guarantees are central and this does not constitute a limitation. 

Next, we introduce three baselines and then, compare their privacy guarantees with those of the proposed scheme. Note that the previous works on random participation provide client-level pure DP guarantees. Hence, comparing these schemes with our work would not be fair. Nevertheless, in \cref{sec:previous_work_numerical}, we provide some comparisons to emphasize what improvement our scheme brings upon previous work.

\subsection{Baselines}

\subsubsection{\OnlyLocal (\OnlyLocalAcr)}
The first setting we consider is the one in which all the clients participate in each iteration, i.e., $p=1$, and each client samples from its local dataset such that each element is sampled with probability $q$ in an i.i.d. fashion. We consider \OnlyLocalAcr to demonstrate the benefits of random participation. The privacy guarantees of local subsampling in \OnlyLocalAcr is given by the following theorem.

\begin{theorem}\label{thm:only_local}
Each iteration of \OnlyLocalAcr scheme is $(\varepsilon,\delta)$-DP, where, for any $\varepsilon'>0$,
\begin{equation}
    \varepsilon=\log\left(1+q\left(e^{\varepsilon'-1}\right)\right),
\end{equation}

\begin{equation}
\delta = q\left(\Phi\left(\frac{C}{2\sigma}-\frac{\sigma \varepsilon'}{C}\right)-e^{\varepsilon'}\Phi\left(-\frac{C}{2\sigma}-\frac{\sigma \varepsilon'}{C}\right)\right).
\end{equation}
\end{theorem}

The proof of \cref{thm:only_local} is based on \cref{thm:only-local-sampling} and given in \cref{sec:proofs}.

\subsubsection{\UpperBound (\UpperBoundAcr)}
In this setting, our sampling procedure is exactly the same as in \cref{alg:DP-DSGD} but we assume that when the clients randomly decide whether to participate or not, the identities of the participating clients are disclosed. Hence, in \UpperBoundAcr, random participation of clients helps amplifying DP guarantees only since when some clients are not sampled, no information is leaked from them. This obviously results in weaker privacy guarantees than those of \cref{alg:DP-DSGD}, and we consider this setting as an upper bound. To elaborate more, let us assume $D'$ has exactly the same elements as $D$ except an additional element $x'$. If the client storing $x'$ is sampled, then the privacy leakage is the same as \OnlyLocalAcr. On the other hand, when this specific client is not sampled, no information is leaked at all about the existence of $x'$. This implies that the privacy guarantees of \UpperBoundAcr should still be better than \OnlyLocalAcr.  In the following theorem, we present the formal privacy guarantees of \UpperBoundAcr.

\begin{theorem}\label{thm:upper_bound}
Each iteration of \UpperBoundAcr scheme is $(\varepsilon,\delta)$-DP, where, for any $\varepsilon'>0$

\begin{equation}
    \varepsilon = \log\left(1+pq\left(e^{\varepsilon'}-1\right)\right),
\end{equation}

\begin{equation}
\delta = pq\left(\Phi\left(\frac{C}{2\sigma}-\frac{\sigma \varepsilon''}{C}\right)-e^{\varepsilon''}\Phi\left(-\frac{C}{2\sigma}-\frac{\sigma \varepsilon''}{C}\right)\right),
\end{equation}
where $\varepsilon'' = \varepsilon' + \log\left(e^{\varepsilon-\varepsilon'} + (1-e^{\varepsilon-\varepsilon'})\frac{p(1-q)}{1-pq}\right)$.
\end{theorem}

The proof of \cref{thm:upper_bound} is given in \cref{sec:proofs}.

\subsubsection{\LowerBound (\LowerBoundAcr)} In \LowerBoundAcr, we have the same client sampling and dataset sampling procedure with probabilities $p$ and $q$, respectively, but at the beginning of each iteration, the elements stored in all the clients are shuffled centrally and uniformly. This is obviously too costly and far from practice, but we consider this case as a lower bound to measure the performance of our scheme. The privacy guarantees of \LowerBoundAcr are derived in the same way as in \cref{thm:only_local}. Notice that the case in \cref{thm:only_local} and \LowerBoundAcr are exactly the same except for the probability of an element being sampled, which is $pq$ in \LowerBoundAcr and $q$ in \OnlyLocalAcr. Hence, the next corollary follows.

\begin{corollary}
Each iteration of the \LowerBoundAcr scheme is $(\varepsilon,\delta)$-DP where, for any $\varepsilon'>0$,
\begin{equation}
    \varepsilon=\log\left(1+q\left(e^{\varepsilon'-1}\right)\right),
\end{equation}

\begin{equation}
\delta = q\left(\Phi\left(\frac{C}{2\sigma}-\frac{\sigma \varepsilon'}{C}\right)-e^{\varepsilon'}\Phi\left(-\frac{C}{2\sigma}-\frac{\sigma \varepsilon'}{C}\right)\right).
\end{equation}
\end{corollary}

\subsection{Comparison to Baselines}
\label{sec:numerical_results}
\begin{figure}[t]
\begin{small}
\centering
\resizebox{0.6\linewidth}{!}{
\input{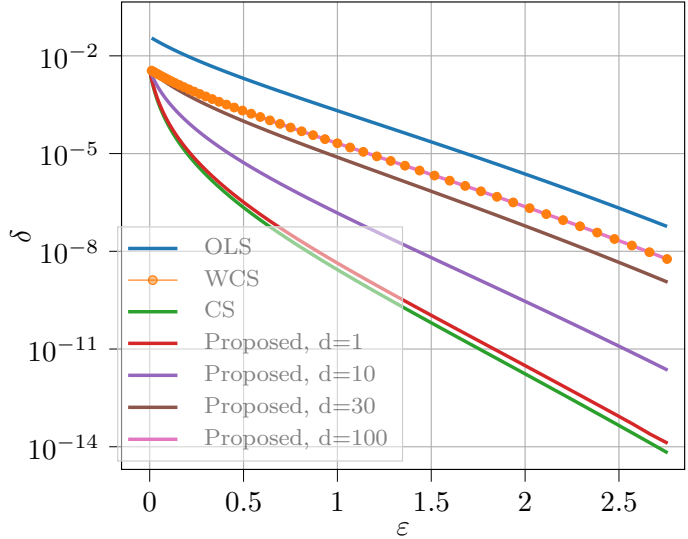}}
\caption{$\delta$ vs $\epsilon$ trade-off for the considered schemes.\label{fig:eps_vs_delta}}
\end{small}
\end{figure}

\begin{figure}[t]
\begin{small}
\centering
\resizebox{0.6\linewidth}{!}{
\begin{tikzpicture}

\definecolor{color0}{rgb}{0.12156862745098,0.466666666666667,0.705882352941177}
\definecolor{color1}{rgb}{1,0.498039215686275,0.0549019607843137}
\definecolor{color2}{rgb}{0.172549019607843,0.627450980392157,0.172549019607843}
\definecolor{color3}{rgb}{0.83921568627451,0.152941176470588,0.156862745098039}
\definecolor{color4}{rgb}{0.580392156862745,0.403921568627451,0.741176470588235}
\definecolor{color5}{rgb}{0.549019607843137,0.337254901960784,0.294117647058824}
\definecolor{color6}{rgb}{0.890196078431372,0.466666666666667,0.76078431372549}

\begin{axis}[
legend cell align={left},
legend style={fill opacity=0.8, draw opacity=1, text opacity=1, draw=white!80!black},
log basis x={10},
tick align=outside,
tick pos=left,
x grid style={white!69.0196078431373!black},
xlabel={$\sigma$},
xmin=0.794328234724281, xmax=12, 
xmode=log,
xtick style={color=black},
y grid style={white!69.0196078431373!black},
ylabel={$\varepsilon$},
ymin=-0.08412145240983, ymax=2.78,
ytick style={color=black},
grid
]

\addplot [very thick, color0]
table {%
1 2.18169424950752
1.27427498570313 1.29414973441552
1.62377673918872 0.760829226776709
2.06913808111479 0.458283422109959
2.63665089873036 0.287563364326514
3.35981828628378 0.188210087083747
4.28133239871939 0.127755556949025
5.45559478116852 0.0892879371844971
6.95192796177561 0.0638279217748418
8.85866790410082 0.0464184252664385
11.2883789168469 0.034197914741315
};
\addlegendentry{\scriptsize OLS}
\addplot [very thick, color1, mark=*, mark size=2, mark options={solid}]
table {%
1 1.68454391323194
1.27427498570313 0.972857313233232
1.62377673918872 0.568299862110192
2.06913808111479 0.345055775105299
2.63665089873036 0.219334826302288
3.35981828628378 0.145321725612098
4.28133239871939 0.0995720570177998
5.45559478116852 0.0700152184962807
6.95192796177561 0.0502020050494863
8.85866790410082 0.0365209333789288
11.2883789168469 0.0268522788408023
};
\addlegendentry{\scriptsize \UpperBoundAcr}


\addplot [very thick, color5]
table {%
1 1.68433433568776
1.17210229753348 1.17621960391201
1.37382379588326 0.82077006289945
1.61026202756094 0.57840934530768
1.8873918221351 0.414477792765665
2.21221629107045 0.302841551780755
2.59294379740467 0.225567838805205
3.0391953823132 0.170923046657799
3.56224789026244 0.131342221308386
4.1753189365604 0.101916344588635
4.89390091847749 0.0794056730055682
5.73615251044868 0.0616532787291906
6.72335753649934 0.0471737903521332
7.88046281566991 0.0315014493266203
9.23670857187386 0.018903487212238
10.8263673387405 0.0108363890470265
};
\addlegendentry{\scriptsize d=100}

\addplot [very thick, color4]
table {%
1 1.42502410766373
1.17210229753348 0.95457596804233
1.37382379588326 0.636629393076201
1.61026202756094 0.426522186934014
1.8873918221351 0.288133782388438
2.21221629107045 0.196179598244069
2.59294379740467 0.134301271171625
3.0391953823132 0.0922209825487075
3.56224789026244 0.063443913149025
4.1753189365604 0.04374953599955
4.89390091847749 0.030298772217418
5.73615251044868 0.0211323757960657
6.72335753649934 0.0147361582068907
7.88046281566991 0.0092474662291403
9.23670857187386 0.0057686070536668
10.8263673387405 0.00372239087176752
};
\addlegendentry{\scriptsize d=30}

\addplot [very thick, color3]
table {%
1 0.720210096221418
1.17210229753348 0.427973291709783
1.37382379588326 0.261005305302713
1.61026202756094 0.164688554583359
1.8873918221351 0.107408305979869
2.21221629107045 0.0721052923221486
2.59294379740467 0.0496110263990982
3.0391953823132 0.0348637462227017
3.56224789026244 0.0249613967685036
4.1753189365604 0.0181753383318714
4.89390091847749 0.0134405033304452
5.73615251044868 0.010082121219824
6.72335753649934 0.00764939958017451
7.88046281566991 0.00539586679779992
9.23670857187386 0.00373768121786054
10.8263673387405 0.00263581508626791
};
\addlegendentry{\scriptsize d=10}

\addplot [very thick, color2]
table {%
1 0.393456480052888
1.17210229753348 0.219964888965301
1.37382379588326 0.1304577813718
1.61026202756094 0.0820761852602339
1.8873918221351 0.0543693775951229
2.21221629107045 0.0375861930294409
2.59294379740467 0.026899167299159
3.0391953823132 0.0197961879514486
3.56224789026244 0.014901408085234
4.1753189365604 0.01142420652766
4.89390091847749 0.00889008087442262
5.73615251044868 0.00700304100423198
6.72335753649934 0.00557197428500613
7.88046281566991 0.00419418202023675
9.23670857187386 0.00306518186408194
10.8263673387405 0.00225653673074186
};
\addlegendentry{\scriptsize d=1}

\end{axis}

\end{tikzpicture}}
\caption{$\varepsilon$ vs $\sigma$ trade-off for the considered schemes.\label{fig:eps_vs_sigma}}
\end{small}
\end{figure}
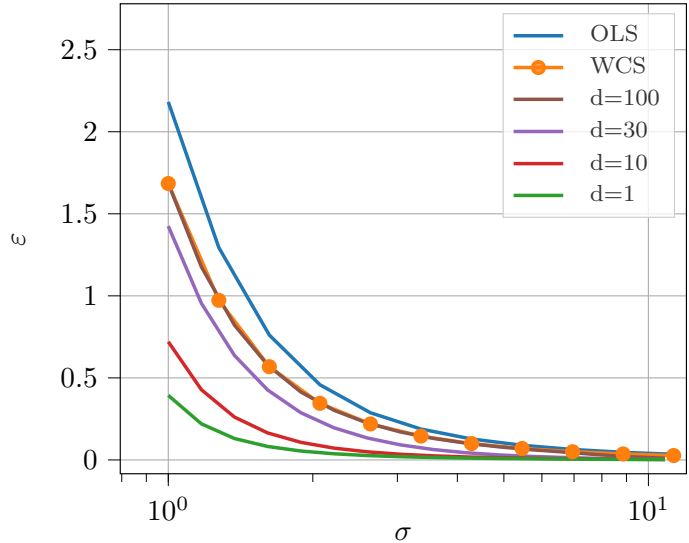

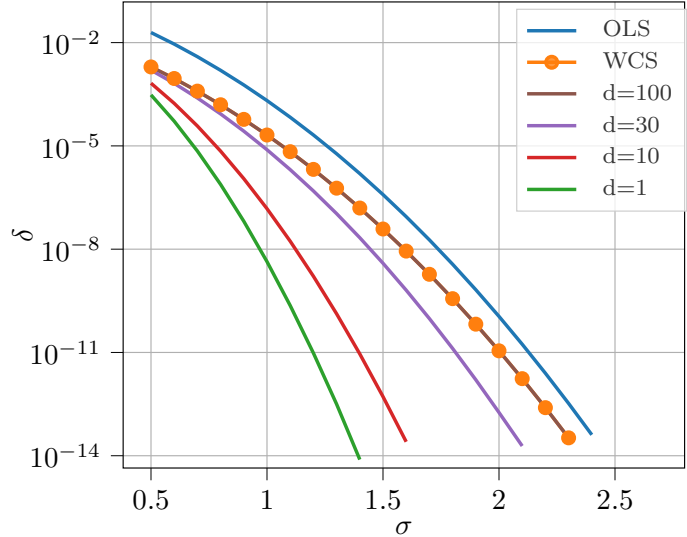
\begin{figure}[t]
\begin{small}
\centering
\resizebox{0.6\linewidth}{!}{
\begin{tikzpicture}

\definecolor{color0}{rgb}{0.12156862745098,0.466666666666667,0.705882352941177}
\definecolor{color1}{rgb}{1,0.498039215686275,0.0549019607843137}
\definecolor{color2}{rgb}{0.172549019607843,0.627450980392157,0.172549019607843}
\definecolor{color3}{rgb}{0.83921568627451,0.152941176470588,0.156862745098039}
\definecolor{color4}{rgb}{0.580392156862745,0.403921568627451,0.741176470588235}
\definecolor{color5}{rgb}{0.549019607843137,0.337254901960784,0.294117647058824}
\definecolor{color6}{rgb}{0.890196078431372,0.466666666666667,0.76078431372549}

\begin{axis}[
legend cell align={left},
legend style={
  fill opacity=0.8,
  draw opacity=1,
  text opacity=1,
  at={(0.70,0.55)},
  anchor=south west,
  draw=white!80!black
},
log basis y={10},
tick align=outside,
tick pos=left,
x grid style={white!69.0196078431373!black},
xlabel={$\sigma$},
xmin=0.38, xmax=2.8,
xtick style={color=black},
y grid style={white!69.0196078431373!black},
ylabel={$\delta$},
ymin=4.39716289573844e-15, ymax=0.15185104230398,
ymode=log,
ytick style={color=black},
grid
]
\addplot [very thick, color0]
table {%
0.5 0.019647881116579
0.6 0.00907191234023005
0.7 0.00390022034626616
0.8 0.00156765892809007
0.9 0.000589402397112209
1 0.000207121486021392
1.1 6.79442210176737e-05
1.2 2.07784919706791e-05
1.3 5.91630405372075e-06
1.4 1.56655967636957e-06
1.5 3.85335668676559e-07
1.6 8.79656541820433e-08
1.7 1.86209088987408e-08
1.8 3.6523817371215e-09
1.9 6.63360322543957e-10
2 1.11497105742152e-10
2.1 1.73336240108206e-11
2.2 2.49127869590351e-12
2.3 3.30886750320628e-13
2.4 4.05972190154945e-14
};
\addlegendentry{\scriptsize OLS}
\addplot [very thick, color1, mark=*, mark size=2, mark options={solid}]
table {%
0.5 0.0019647881116579
0.6 0.000907191234023005
0.7 0.000390022034626616
0.8 0.000156765892809007
0.9 5.89402397112209e-05
1 2.07121486021392e-05
1.1 6.79442210176737e-06
1.2 2.07784919706791e-06
1.3 5.91630405372075e-07
1.4 1.56655967636957e-07
1.5 3.85335668676559e-08
1.6 8.79656541820433e-09
1.7 1.86209088987408e-09
1.8 3.6523817371215e-10
1.9 6.63360322543958e-11
2 1.11497105742152e-11
2.1 1.73336240108206e-12
2.2 2.49127869590351e-13
2.3 3.30886750320628e-14
};
\addlegendentry{\scriptsize \UpperBoundAcr}

\addplot [very thick, color5]
table {%
0.5 0.00196449299758502
0.6 0.000906979482991872
0.7 0.000389890767358061
0.8 0.000156693389468536
0.9 5.89041013707004e-05
1 2.06957927422025e-05
1.1 6.78767653720769e-06
1.2 2.07530817220913e-06
1.3 5.90754713449826e-07
1.4 1.56379552684882e-07
1.5 3.84535793784835e-08
1.6 8.77533313570211e-09
1.7 1.85691948928479e-09
1.8 3.64084087323136e-10
1.9 6.61019328163093e-11
2 1.11497105742152e-11
2.1 1.73336240108206e-12
2.2 2.49127869590351e-13
2.3 3.30886750320628e-14
};
\addlegendentry{\scriptsize d=100}

\addplot [very thick, color4]
table {%
0.5 0.00158903337173989
0.6 0.000655234721158955
0.7 0.000246554775076113
0.8 8.49553562914451e-05
0.9 2.68081642778739e-05
1 7.73920906098624e-06
1.1 2.04125853965564e-06
1.2 4.91226298571235e-07
1.3 1.07719619890645e-07
1.4 2.14999075964784e-08
1.5 3.90174818676314e-09
1.6 6.43228865726542e-10
1.7 9.625114927303e-11
1.8 1.3064749282421e-11
1.9 1.6082779552562e-12
2 1.80193637788761e-13
2.1 1.91846538655227e-14
};
\addlegendentry{\scriptsize d=30}

\addplot [very thick, color3]
table {%
0.5 0.000662257204619827
0.6 0.000172120648120284
0.7 3.7811119018869e-05
0.8 7.05155754094022e-06
0.9 1.11659829926225e-06
1 1.49933738633479e-07
1.1 1.70434606161507e-08
1.2 1.63729573898763e-09
1.3 1.32712791867107e-10
1.4 9.06343444739832e-12
1.5 5.20969933859306e-13
1.6 2.54374299402116e-14
};
\addlegendentry{\scriptsize d=10}

\addplot [very thick, color2]
table {%
0.5 0.000303111878051041
0.6 5.22209594782908e-05
0.7 7.08574850857957e-06
0.8 7.609407981235e-07
0.9 6.46859942321498e-08
1 4.34637794199944e-09
1.1 2.30390639899269e-10
1.2 9.61549062594713e-12
1.3 3.15303338993545e-13
1.4 7.74491581978509e-15
1.5 -3.5527136788005e-17
1.6 -1.06581410364015e-16
1.7 -1.06581410364015e-16
1.8 -1.06581410364015e-16
1.9 -1.06581410364015e-16
2 -1.06581410364015e-16
2.1 -1.06581410364015e-16
2.2 -1.06581410364015e-16
2.3 -1.06581410364015e-16
2.4 -1.06581410364015e-16
2.5 -1.06581410364015e-16
2.6 -1.06581410364015e-16
2.7 -1.06581410364015e-16
2.8 -1.06581410364015e-16
2.9 -1.06581410364015e-16
};
\addlegendentry{\scriptsize d=1}

\end{axis}

\end{tikzpicture}}
\caption{$\delta$ vs $\sigma$ trade-off for the considered schemes.\label{fig:delta_vs_sigma}}
\end{small}
\end{figure}

For a single iteration of the learning task, we present the trade-offs $\varepsilon$ vs. $\delta$, $\varepsilon$ vs. $\sigma$ and $\delta$ vs. $\sigma$, in \cref{fig:eps_vs_delta}, \cref{fig:eps_vs_sigma} and \cref{fig:delta_vs_sigma}, respectively, for the proposed scheme and the baselines we consider. While generating the plots, we assume $C=1$ and set $\sigma=1$, $p=q=0.1$, unless otherwise stated. We observe that the proposed scheme's privacy guarantees lie in between \LowerBoundAcr and \UpperBoundAcr. Inferring whether a client is sampled or not by observing the corresponding model update is easier if the participation of one client changes the model update significantly, i.e., large sensitivity to addition or removal of one client. Since the sensitivity of the gradient sum received by the PS with respect to one element is bounded by $C$, the number of sampled elements from a client $i$, i.e., $|S_{i,t}|$, determines how easy it is to make an inference about the participation of that client. That is because the sensitivity of the gradient sum with respect to the participation of client $i$ becomes $|S_{i,t}|C$. Since smaller $d\cdot q$ values implies smaller expected values of $|S_{i,t}|C$, in the figures, we observe that smaller $d\cdot q$ values result in stronger privacy guarantees, and our scheme performs quite close to the \LowerBoundAcr when $d\cdot q$ is small. For example, for $d=1$, \LowerBoundAcr and our scheme are almost identical since it is quite hard to distinguish if the sole element of each client is sampled or not and hence, it is hard to make inference about the participation of any client. Increasing $d$ gives more information to an adversary about the identities of the sampled clients, but for $d=10$ the privacy guarantees if our scheme and still very close to \LowerBoundAcr. As the value of $d\cdot q$ increases, so does the privacy leakage of our scheme as we observe in all three figures, and the privacy guarantees of the proposed scheme converge to those of \UpperBoundAcr as $d\to \infty$. For the considered set of parameters, since there are no significant differences between the privacy guarantees of \UpperBoundAcr and the proposed scheme for values $d\cdot q>10$, for clarity, we only show $d=100$, which is almost the same curve as \UpperBoundAcr. Such a convergence of our scheme to \UpperBoundAcr is intuitive since when $d$ is large, an adversary can gain a significant amount of information about the number of sampled clients and their identities. Since the assumption in \UpperBoundAcr is that the identities of the sampled clients are known, our scheme becomes very close to the \UpperBoundAcr, when $d$ is large.

Note that in many practical cases, the dataset sizes may be much larger than those considered in this section. This does not necessarily mean that for the cases $d>100$, the proposed scheme will be always equivalent to \UpperBoundAcr. If the local dataset sizes of the clients are large, then much smaller $q$ values can be employed. In fact, smaller $d\cdot q$ values result in better privacy guarantees in general. This is the essence of what we have observed in this section. We consider an example with larger datasets in \cref{subsec:large_datasets}. 

\subsection{Comparison to Prior Work} \label{sec:previous_work_numerical}

\begin{figure}[ht]
\begin{small}
\centering
\resizebox{0.5\linewidth}{!}{
\input{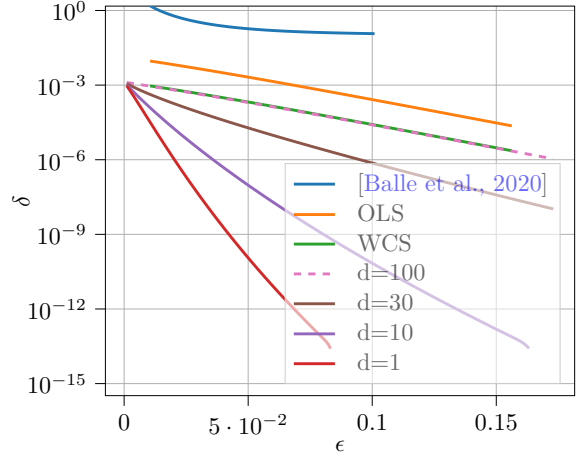}}
\caption{Comparison between \cite{balle2020privacy} and our baselines in terms of $\varepsilon$ vs. $\delta$ trade-off. \label{fig:comparison_with_literature}}
\end{small}
\end{figure}

Previous works on random participation provide client-level guarantees and assumes pure $\varepsilon$-DP guarantees. Hence, we think that the comparison of these schemes with our work would not be fair for these works since the conversion from $\varepsilon$-DP guarantees to $(\varepsilon,\delta)$-DP considerably degrades the privacy guarantees. Further, these guarantees are client-level, so they may not be directly comparable with our results. Nevertheless, here we provide the comparison of our results with those of \cite{balle2020privacy} in \cref{fig:comparison_with_literature} for a region that we were able to get meaningful privacy guarantees for \cite{balle2020privacy}. We used sampling probabilities $p=q=0.1$,
the sensitity per gradient $C= 1.0$, noise std $\sigma = 3.0$. As we observe, in terms of approximate DP guarantees, the proposed scheme significantly improves upon \cite{balle2020privacy}.

\subsection{Local Sampling vs. Client Sampling}\label{sec:local_vs_client}

Due to \cref{thm:main}, when we decrease $p$ and $q$, the DP guarantees of our scheme improves. However, we may have some constraints that limit the minimum values $p$ and $q$ can take. For example, batch size is a crucial hyperparameter affecting the accuracy of learning algorithms, and in most cases, choosing the batch size too small results in too noisy updates while choosing it too large may forbid learning details in the dataset. Hence, the value of $p \cdot q$ should be chosen to attain the required batch size. In this section, we investigate how the DP guarantees of the proposed scheme and our baselines are affected by the choice of the individual values of $p$ and $q$ given that $pq$ is fixed.

\begin{figure}[ht]
\begin{small}
\centering
\resizebox{0.6\linewidth}{!}{
\begin{tikzpicture}

\definecolor{color0}{rgb}{0.12156862745098,0.466666666666667,0.705882352941177}
\definecolor{color1}{rgb}{1,0.498039215686275,0.0549019607843137}
\definecolor{color2}{rgb}{0.172549019607843,0.627450980392157,0.172549019607843}
\definecolor{color3}{rgb}{0.83921568627451,0.152941176470588,0.156862745098039}
\definecolor{color4}{rgb}{0.580392156862745,0.403921568627451,0.741176470588235}

\begin{axis}[
legend cell align={left},
legend style={
  fill opacity=0.8,
  draw opacity=1,
  text opacity=1,
  at={(0.03,0.97)},
  anchor=north west,
  draw=white!80!black
},
log basis x={10},
log basis y={10},
tick align=outside,
tick pos=left,
x grid style={white!69.0196078431373!black},
xlabel={$q$},
xmin=6.30957344480193e-05, xmax=1.58489319246111,
xmode=log,
xtick style={color=black},
y grid style={white!69.0196078431373!black},
ylabel={$\varepsilon$},
ymin=2.06907089701209e-04, ymax=8.80786576934316,
ymode=log,
ytick style={color=black},
grid
]
\addplot [very thick, color3]
table {%
0.0001 0.000914901321981594
0.000193069772888325 0.00233815636214832
0.000372759372031494 0.00581238888623976
0.000719685673001152 0.0141214936657902
0.00138949549437314 0.0335639588225359
0.00268269579527972 0.077731914988613
0.00517947467923121 0.173349937450939
0.01 0.36394787666132
0.0193069772888325 0.698003086543437
0.0372759372031494 1.19562289021589
0.0719685673001151 1.82949452416683
0.138949549437314 2.54934143925593
0.268269579527972 3.31330025567026
0.517947467923121 4.09652151381563
1 4.88655418162691
};
\addlegendentry{\scriptsize OLS}

\addplot [very thick, color4]
table {%
0.0001 0.000914901321981594
0.000193069772888325 0.0017656465192498
0.000372759372031494 0.00340613351463276
0.000719685673001152 0.00656582442139208
0.00138949549437314 0.0126381310916818
0.00268269579527972 0.0242586163773591
0.00517947467923121 0.0463192722921131
0.01 0.0875822114532023
0.0193069772888325 0.162731433141487
0.0372759372031494 0.293560335002808
0.0719685673001151 0.506059887640207
0.138949549437314 0.820586984428772
0.268269579527972 1.23997441089737
0.517947467923121 1.74761191951534
1 2.31778889902168
};
\addlegendentry{\scriptsize \UpperBoundAcr}

\addplot [very thick, color0, mark=x, mark size=3, mark options={solid}]
table {%
0.0001 0.000914901465269273
0.000193069772888325 0.000997132736460039
0.000372759372031494 0.00117654903263441
0.000719685673001152 0.0016144035626929
0.00138949549437314 0.00292944417863286
0.00268269579527972 0.00845318464087523
0.00517947467923121 0.0351786555137455
0.01 0.0871199312307608
0.0193069772888325 0.162731301194846
0.0372759372031494 0.29356039924873
0.0719685673001151 0.506060011466537
0.138949549437314 0.820587172395371
0.268269579527972 1.23997486581875
0.517947467923121 1.74761281909826
1 2.31778889902168
};
\addlegendentry{\scriptsize d=1000}

\addplot [very thick, color2, mark=x, mark size=3, mark options={solid}]
table {%
0.0001 0.000914901465269273
0.000193069772888325 0.000917383662899266
0.000372759372031494 0.000922195729474924
0.000719685673001152 0.000931560255619451
0.00138949549437314 0.000949919149068168
0.00268269579527972 0.000986429141132271
0.00517947467923121 0.00106107528648491
0.01 0.00122214825715093
0.0193069772888325 0.00160849747024662
0.0372759372031494 0.00275239590451638
0.0719685673001151 0.00795952212048913
0.138949549437314 0.0656740011648294
0.268269579527972 1.0598515334096
0.517947467923121 1.74760931603292
1 2.31778889902168
};
\addlegendentry{\scriptsize d=30}
\addplot [very thick, color1, mark=x, mark size=3, mark options={solid}]
table {%
0.0001 0.000914901465269273
0.000193069772888325 0.000915066961112182
0.000372759372031494 0.000915386610593258
0.000719685673001152 0.000916004231216575
0.00138949549437314 0.000917198439923385
0.00268269579527972 0.000919510713505104
0.00517947467923121 0.000923999825783665
0.01 0.000932760473623809
0.0193069772888325 0.000950030953117441
0.0372759372031494 0.000984760918517416
0.0719685673001151 0.00105743180855182
0.138949549437314 0.00122253241696745
0.268269579527972 0.00167339919666583
0.517947467923121 0.00370777844209658
1 2.31778889902168
};
\addlegendentry{\scriptsize d=2}

\end{axis}

\end{tikzpicture}}
\caption{Trade-off between $q$ and $\varepsilon$ when $pq=10^{-4}$. \label{fig:p_q_trade_off}}
\end{small}
\end{figure}
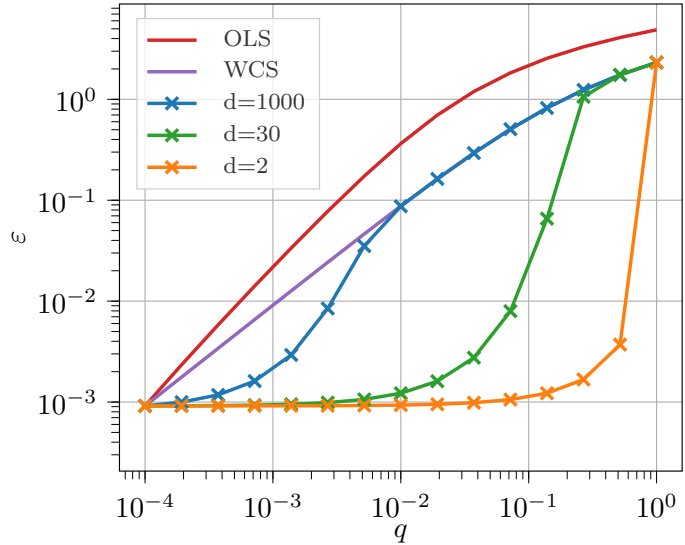

In \cref{fig:p_q_trade_off}, given $pq=10^{-4}$, we plot the $\varepsilon$ vs. $q$ trade-off, when $C=1$ and $\sigma=1$. We observe that, while we increase $q$, which corresponds to decreasing $p$ at the same rate, the DP guarantees of all the schemes degrade. When all clients participate, i.e., $p=1.0$, all the schemes become equivalent and $\varepsilon$ is at its best value. Although this implies that we should rely only on local sampling to meet our target batch size, in most real world scenarios, this is not possible for the reasons we have discussed in \cref{sec:motivation}, and DP guarantees are dictated by \cref{thm:main}.

In accordance with the observations in \cref{sec:numerical_results}, we further observe that as $q$ increases, smaller dataset sizes at the clients, i.e., smaller $d$, become more favorable since the proposed scheme converges to \UpperBoundAcr at larger values of $q$. 

\subsection{Modification of the proposed scheme when the PS is not trusted} \label{subsec:untrusted_ps}

When the PS is not trusted, client sampling can still be applied, and our theoretical analysis is still valid. In \cref{alg:DP-DSGD}, since Gaussian noise is added by the PS, the same noise protects all the clients' updates at the same time. In case of an untrusted PS, we cannot rely on the noise addition by the PS; and hence, each client must add Gaussian noise to protect its own privacy. This causes a higher variance of the effective noise added to the total update received by the PS; and hence, reduces the accuracy compared to the trusted PS case. 

On the other hand, the sampling can still be done in the following manner. First, each client decides to participate or not independently with probability $p$, and if a client decides not to participate, since we do not want the PS to learn if a client is sampled or not, it sends pure noise to the PS without processing its local dataset. If it decides to participate, then it samples its local dataset, calculates the gradients of the sampled elements, and sends the average of the gradients by adding Gaussian noise. This way, the PS will always receive a message from all the available clients, and our analysis on client sampling is still valid. One disadvantage of this modified protocol is that an available client sends a message to the PS at each iteration, even when it is not participating, and hence, the communication efficiency of the scheme degrades, which is one of the main motivations of the client sampling. Note that the reduction in the computational costs due to client sampling is still valid.

If the communication costs are scarce in a setting, alternatively, we may prefer a client communicates via the PS only if it decides to participate in a round. In this case, the PS would clearly learn if a client is participating or not, and our privacy guarantees reduce to the one provided by \UpperBoundAcr. Although it provides weaker guarantees than the proposed method, we have seen in \cref{sec:discussion_and_numerical} and \cref{sec:empirical_eval} that \UpperBoundAcr considerably improves upon \OnlyLocalAcr, and can still provide meaningful privacy guarantees.

\section{Empirical Evaluation}
\label{sec:empirical_eval}
In this section, we consider two different experimental settings to empirically show the improvement on the accuracy when the variance of the Gaussian noise added for DP guarantees is determined according to \cref{thm:main}. In both settings, we use EMNIST dataset \cite{cohen2017emnist}, which is an extension of the MNIST dataset \cite{lecun1998gradient}, where, in addition to the handwritten digits, uppercase and lowercase letters are also included. Although there are different possible splits in this dataset, we use the split based on digit and letter classification, referred to as \emph{ByClass} in \cite{cohen2017emnist}. Hence, we have 62 possible classes. This dataset includes 814255 data samples in total, and we use 697932 of them as the training set and the rest as the test set, as implemented in PyTorch \cite{pytorch}. In both scenarios, we use a simple deep neural network, which is composed of two layers of CNNs followed by two fully connected layers. We give further details on the experimental setting in \cref{sec:experiment_details}. We also provide the code for the experiments in [github link]\footnote{Please see the supplementary material for the code}.

\subsection{Many Clients and Small Datasets Scenario}\label{subsec:small_datasets}
In the first scenario, we assume there are large number of users while each user has only a few number of samples. Motivated by the findings in \cref{fig:eps_vs_delta}, we choose that each client has $d=30$ samples and we equally split the training set across 23264 clients. Since we have relatively large number of clients, we choose the client sampling probability as $p=0.001$, and each client that decides to participate samples its local dataset with probability $q=0.1$. These sampling rates correspond to an average batch size of 70, which we verify to provide a good accuracy via hyperparameter search. For each iteration of \cref{alg:DP-DSGD}, we require a fixed privacy guarantee such that $\varepsilon=0.015$ and $\delta=10^{-6}$. These values may seem too conservative, but note that training deep learning models is an iterative procedure, and at the end of training, the total privacy leakage should be reasonably low. To attain these guarantees, for the proposed method, the server should add Gaussian noise with a standard deviation of $\sigma=1.065$ according to \cref{thm:main}. Similarly, for \UpperBoundAcr, according to \cref{thm:upper_bound}, we should have $\sigma=7.65$, and for
\OnlyLocalAcr, according \cref{thm:only_local}, we should have $\sigma=22.4$. Moreover, to provide a baseline, we also train our model ignoring privacy constraints, i.e., $\sigma=0$. For the proposed method, \UpperBoundAcr and \OnlyLocalAcr, we use a gradient clipping value of $C=1$, while no such restriction is imposed in the non-private case.

\begin{figure}[t]
\begin{small}
\centering
\resizebox{0.6\linewidth}{!}{
\input{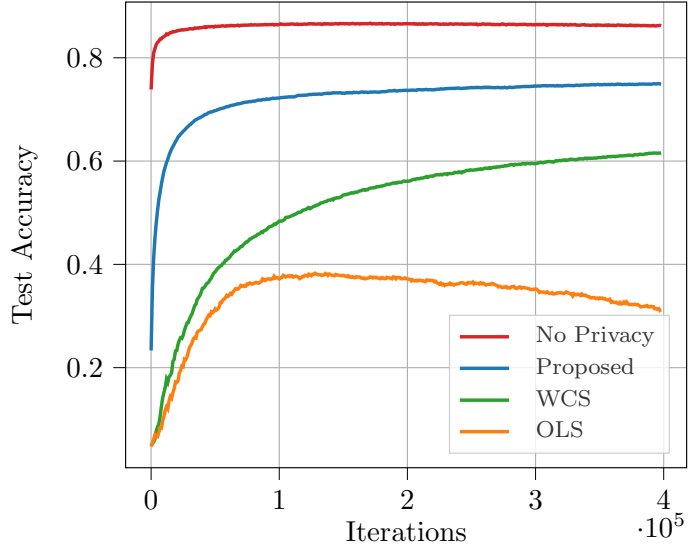}}
\caption{Many clients and small datasets scenario.\label{fig:experiments_smallds}}
\end{small}
\end{figure}

In \cref{fig:experiments_smallds}, we present the test accuracies of the proposed scheme and all the baselines with respect to the number of iterations. We observe that the proposed method attains much higher accuracy than the other analysis methods.  We further observe that, although it constitutes an upper bound on $(\varepsilon,\delta)$, \UpperBoundAcr still brings a non-trivial improvement over \OnlyLocalAcr. In fact, this shows that even if the identities of the sampled clients can be inferred with high confidence from the gradients, analysis provided in \cref{thm:upper_bound} for client sampling still provides useful privacy guarantees. Finally, we observe that the analysis based on only local sampling, i.e., \OnlyLocalAcr, has quite a poor performance due to the very high noise addition in this scheme, in order to attain $\varepsilon=0.015$.

\subsection{Few Clients and Large Datasets Scenario}\label{subsec:large_datasets}

Next, we consider the scenario with fewer clients, but each client has a larger dataset. We assume each client stores $d=1000$ data points and there are 697 clients. Each client is independently sampled with a probability of $p=0.1$ and each participating client samples its data points independently each with probability of $q=0.001$. Similar to the scenario in \cref{subsec:small_datasets}, we have an average batch size of 70. Again, we fix per-iteration privacy guarantees as $\varepsilon=0.015$ and $\delta=10^{-6}$. To attain them, according to \cref{thm:main}, the server should add Gaussian noise with $\sigma=0.646$ for the proposed scheme. For \UpperBoundAcr and \OnlyLocalAcr, we have $\sigma=0.873$ and $\sigma=1.103$, according to \cref{thm:upper_bound} and \cref{thm:only_local}, respectively. While we use $C=1$ for the gradient clipping parameter in the private cases, we do not clip the gradients or add any noise in the non-private case.

\begin{figure}[ht]
\begin{small}
\centering
\resizebox{0.6\linewidth}{!}{
\input{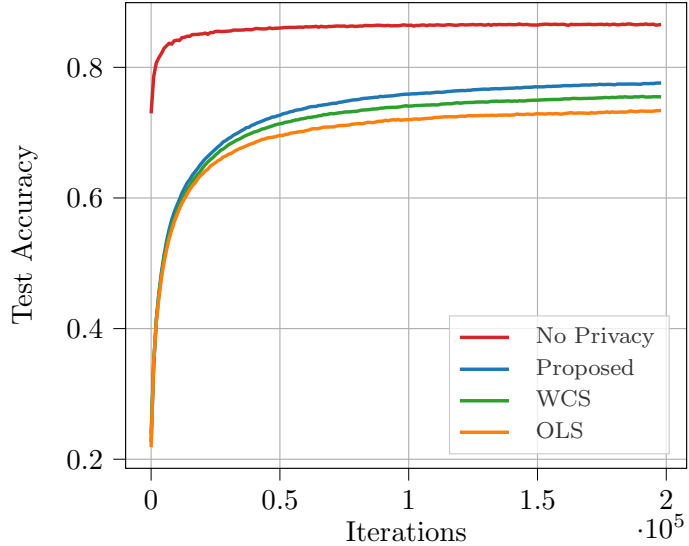}}
\caption{Few clients and large datasets scenario.\label{fig:experiments_largeds}}
\end{small}
\end{figure}
In \cref{fig:experiments_largeds}, we present our experiment results. Again, we observe that the accuracy of each scheme is inversely proportional to the amount of added noise; and hence, the proposed scheme has the best accuracy. However, compared to the setting in \cref{subsec:small_datasets}, in this setting, the performance of the schemes are closer to each other. As discussed in \cref{sec:local_vs_client}, that is because we have a smaller value of $q$ although $pq$ value is the same. Hence, smaller amount of noise is enough to attain $\varepsilon=0.015$.

\section{Conclusion}
Client sampling is widely accepted as a core aspect of FL to reduce the energy consumption of clients and to limit the communication load. In this paper, we have analyzed the privacy amplification provided by client sampling in FL to amplify sample-level DP guarantees. While the privacy amplification effect of local data sampling has been well studied, to the best of our knowledge, this is the first work to study the approximate DP implications of client sampling in FL. Previous works employing client sampling either provide user-level DP guarantees, or their analyses rely on shuffling. Although shuffling provides strong central privacy guarantees when local randomizers with pure DP guarantees are available, when the local randomizers satisfy approximate DP, the central privacy guarantees degrade considerably. This poses a challenge when Gaussian noise is used as a randomizer in FL settings. Instead, we have provided an analysis that does not rely on shuffling and allows the use of Gaussian noise without degrading the privacy guarantees. We have also shown that when the number of samples available at each client is small, we obtain privacy guarantees close to those of centralized training, in which uniform sampling is possible. Moreover, even if the local dataset sizes are very large, we have shown that client sampling is still quite beneficial compared to only local dataset sampling. Moreover, since available clients locally decide to participate or not, independently from each other and from any other system parameter, they can keep track of their own privacy losses; and therefore, our scheme trivially scales to large systems, and is feasible to apply in practice.

\newpage
\bibliography{refs}
\bibliographystyle{apalike}

\newpage
\appendix
\onecolumn

\section{Proofs}\label{sec:proofs}
\subsection{Proof of \cref{thm:main}}
Without loss of generality, we assume that $D'$ has one more element than $D$ and this extra element is stored by the first client, i.e., $D_{1}'=D_{1}\cup\{x'\}$. Thus, the  distribution of the model update in a single iteration when
user 1 stores $D_{1}'$ is 
\begin{equation}
\xi(z)=\left(1-p\right)\xi_{0}(z)+p\left(\left(1-q\right)\xi_{1}(z)+q\xi_{2}(z)\right)\label{eq:xi}
\end{equation}
and the distribution when user 1 stores $D_{1}$ is 
\begin{equation}
\xi'(z)=\left(1-p\right)\xi_{0}(z)+p\xi_{1}(z)\label{eq:xi_prime}
\end{equation}
where $\xi_{0}$ is the distribution when user 1 is not sampled, $\xi_{1}$
is the distribution when user 1 is sampled but $x'$ is not sampled, $\xi_{2}$ is the distribution when both
user 1 and $x'$ are sampled and $z$ is the observed model update, and hence, $z\in \mathbb{R}^m$. In the proposed algorithm,
all these distributions are multivariate Gaussian mixture distributions of dimension $m$. For clearer notation, we omit $z$ in the rest of the section.

Observe that the event $x'$ being sampled has a probability of $pq$.
Thus, we can write \cref{eq:xi} and \cref{eq:xi_prime} as 
\begin{equation}
\xi=\left(1-pq\right)\left(c_{1}\xi_{0}+c_{2}\xi_{1}\right)+pq\xi_{2},\label{eq:xi-pq}
\end{equation}
 and 
\begin{equation}
\xi'=\left(1-pq\right)\left(c_{1}\xi_{0}+c_{2}\xi_{1}\right)+pq\xi_{1},\label{eq:xi_prime-pq}
\end{equation}
where $c_{1}=\frac{(1-p)}{(1-pq)}$ and $c_{2}=\frac{p(1-q)}{(1-pq)}$.

When the output distribution
of a mechanism is a mixture distribution, \emph{advanced joint convexity}
of hockey stick divergence \cite{balle2018privacy}, which we introduce next, helps tightly upper
bounding it.
\begin{theorem}[Theorem 2 in \cite{balle2018privacy}]
\label{thm:ajc}Let $\mu$ and $\mu'$ be measures composed of mixtures
of other measures $\mu_0$, $\mu_1$ and $\mu_1'$ such that $\mu=(1-\gamma)\mu_{0}+\gamma\mu_{1}$
and $\mu'=(1-\gamma)\mu_{0}+\gamma\mu_{1}'$. For $\alpha\geq1$,
we have 
\begin{equation}
\label{eq:advanced_joint_convexity}
D_{\alpha'}\left(\mu||\mu'\right)=\gamma D_{\alpha}\left(\mu_{1}||(1-\beta)\mu_{0}+\beta\mu_{1}'\right)
\end{equation}

where $\alpha'=1+\gamma(\alpha-1)$ and $\beta=\alpha'/\alpha$. The relation in \cref{eq:advanced_joint_convexity}
is referred to as advanced joint convexity of $D_{\alpha}$ in \cite{balle2018privacy}.
\end{theorem}

Using \cref{thm:ajc}, we can write the divergence between these two
measures as follows
\begin{equation}
D_{\alpha}\left(\xi||\xi'\right)=pqD_{\alpha'}\left(\xi_{2}||(1-\beta)\left(c_{1}\xi_{0}+c_{2}\xi_{1}\right)+\beta\xi_{1}\right).\label{eq:ajc_our_case}
\end{equation}
where $\alpha=e^{\varepsilon}$, $\alpha'=e^{\varepsilon'}$, $\beta=e^{\varepsilon-\varepsilon'}$ and $\varepsilon=\log\left(1+pq\left(e^{\varepsilon'}-1\right)\right)$.
Since hockey stick divergence is also jointly convex in its arguments, we can bound
\cref{eq:ajc_our_case} as
\begin{equation}
D_{\alpha}\left(\xi||\xi'\right)\leq pq(1-\beta)c_{1}D_{\alpha'}\left(\xi_{2}||\xi_{0}\right)+pq\left((1-\beta)c_{2}+\beta\right)D_{\alpha'}\left(\xi_{2}||\xi_{1}\right).\label{eq:ajc_ours_loose}
\end{equation}

However, $D_{\alpha'}\left(\xi_{2}||\xi_{0}\right)$ can be quite
large compared to $D_{\alpha'}\left(\xi_{2}||\xi_{1}\right)$ since
$\xi_{2}$ has the effect of all the sampled data points by user 1
while $\xi_{0}$ does not. Thus, \cref{eq:ajc_ours_loose} is not a
tight bound. Instead, we try to bound \cref{eq:ajc_our_case} in a more
involved way. 

Observe that there is a coupling between $\xi_{0},\xi_{1}$ and $\xi_{2}$
such that except sampling of user 1 and $x'$, the
other sampled users and the sampled elements are
the same. Therefore, $\xi_{1}$ has a set of elements sampled from
user 1 except $x'$ and the same sampled users and elements
sampled in $\xi_{0}$. Similarly, $\xi_{2}$ has $x'$ sampled as well as all the users and the
elements sampled in $\xi_{1}$. As we stated
previously, $\xi_{0}, \xi_{1}$ and $\xi_{2}$ are multivariate Gaussian mixture
distributions. Next, let us define $\mathcal{P}$ be the set of all possible samplings of clients given client 1 is not sampled.
Similarly, let us define $\mathcal{S}_{\mathcal{I}}$ as the set of all possible samplings of elements from the clients in the set $\mathcal{I}$ given these clients are sampled. Moreover, let $\mathcal{S}_1$ be the set of all possible samplings of the elements stored in client 1 given $x'$ is not sampled.

Thus, given $P\in \mathcal{P}_t$, $S\in \mathcal{S}_{P}$ and $S_1 \in \mathcal{S}_1$ are actual samplings of clients and the elements, respectively, we can write $\xi_{1}(P,S,S_1)\sim\mathcal{N}(\boldsymbol{\mu}(P\cup\{1\},S\cup S_1),\sigma^{2}\mathbf{I}_m)$, $\xi_{2}(P,S,S_1)\sim\mathcal{N}(\boldsymbol{\mu}(P\cup\{1\},S\cup S_1\cup \{x'\}),\sigma^{2}\mathbf{I}_m)$ and $\xi_0(P,S) \sim \mathcal{N}(\boldsymbol{\mu}(P,S),\sigma^2\mathbf{I}_m)$ where $\boldsymbol{\mu}(P,S)$ is the observed model update when the set of sampled sevices is $P$ and the set of sampled elements is $S$.

As a result, we can write \cref{eq:ajc_our_case}
as 
\begin{multline}
D_{\alpha}\left(\xi||\xi'\right)= pqD_{\alpha'}\Big(\sum_{P\in \mathcal{P}}\sum_{S\in \mathcal{S}_P}\sum_{S_1\in\mathcal{S}_1}\Pr(P,S)\Pr(S_1)\xi_2(P,S,S_1)||
\bar{c_{1}}\sum_{P\in \mathcal{P}}\sum_{S\in \mathcal{S}_P}\Pr(P,S)\xi_0(P,S)\\
+\bar{c_{2}}\sum_{P\in \mathcal{P}}\sum_{S\in \mathcal{S}_P}\sum_{S_1\in\mathcal{S}_1}\Pr(P,S)\Pr(S_1)\xi_1(P,S,S_1)\Big).
\label{eq:divergence_most_general}
\end{multline}
where $\bar{c_{1}}=(1-\beta)c_{1}$ and $\bar{c_{2}}=c_{2}+\beta(1-c_{2})$. Since hockey stick divergence is jointly convex, we can upper bound \cref{eq:divergence_most_general} as
\begin{align}
&D_{\alpha}\left(\xi||\xi'\right)\nonumber\\
&\leq \sum_{P\in \mathcal{P}}\sum_{S\in \mathcal{S}_P}\Pr(P,S) pqD_{\alpha'}\Big(\sum_{S_1\in\mathcal{S}_1}\Pr(S_1)\xi_2(P,S,S_1)||
\bar{c_{1}}\xi_0(P,S)+\bar{c_{2}}\sum_{S_1\in\mathcal{S}_1}\Pr(S_1)\xi_1(P,S,S_1)\Big)\nonumber \\
&\leq \max_{P\in\mathcal{P},S\in \mathcal{S}_P}pqD_{\alpha'}\Big(\sum_{S_1\in\mathcal{S}_1}\Pr(S_1)\xi_2(P,S,S_1)||
\bar{c_{1}}\xi_0(P,S)+\bar{c_{2}}\sum_{S_1\in\mathcal{S}_1}\Pr(S_1)\xi_1(P,S,S_1)\Big).
\label{eq:divergence_most_general_bounded}
\end{align}

Since the hockey stick divergence is shift-invariant, we can take a reference distribution and then define all the other distributions accordingly. Without loss of generality, we take $\xi_0$ as the reference, and assume its mean $\boldsymbol{\mu}=0$. Then, \cref{eq:divergence_most_general_bounded} can be written as 
\begin{multline}
D_{\alpha}\left(\xi||\xi'\right)\leq \max_{P\in\mathcal{P},S\in \mathcal{S}_P}pqD_{\alpha'}\Big(\sum_{S_1\in\mathcal{S}_1}\Pr(S_1)\mathcal{N}(\boldsymbol{\bar{\mu}}_2(S_1),\sigma^2\mathbf{I}_m)||
\bar{c_{1}}\mathcal{N}(0,\sigma^2\mathbf{I}_m)\\
+\bar{c_{2}}\sum_{S_1\in\mathcal{S}_1}\Pr(S_1)\mathcal{N}(\boldsymbol{\bar{\mu}}_1(S_1 ),\sigma^2\mathbf{I}_m)\Big).
\label{eq:divergence_zero_mean}
\end{multline}
where 
\begin{equation}
\boldsymbol{\bar{\mu}}_1(S_1)=\mathcal{N}(\boldsymbol{\mu}(P\cup\{1\},S\cup S_1),\sigma^{2}\mathbf{I}_m)-\mathcal{N}(\boldsymbol{\mu}(P,S),\sigma^2\mathbf{I}_m),
\end{equation}
and 
\begin{equation}
    \boldsymbol{\bar{\mu}}_2(S_2)=\mathcal{N}(\boldsymbol{\mu}(P\cup\{1\},S\cup S_1\cup \{x'\}),\sigma^{2}\mathbf{I}_m)-\mathcal{N}(\boldsymbol{\mu}(P,S),\sigma^2\mathbf{I}_m).
\end{equation}
Observe that we have $||\boldsymbol{\bar{\mu}}_1(S_1)||_2\leq iC$ and $||\boldsymbol{\bar{\mu}}_2(S_1)||_2\leq (i+1)C$ if there are exactly $i$ elements sampled from client 1, i.e., $|S_1|=i$, where $C$ is the sensitivity of the noiseless output to one sample.

Remember from \cref{def:hockey-stick} that the hockey-stick divergence between two distributions is written as
\begin{equation}
D_{\alpha}(\xi||\xi')\triangleq\int_{Z}\left[\xi(z)-\alpha \xi'(z)\right]_{+}d(z).\label{eq:hockey-stick-proof}
\end{equation}
Since all the Gaussian distributions have a diagonal covariance matrices in \cref{eq:divergence_zero_mean}, and $\Pr(S_1)=q^i(1-q)^{d-i}$ when $|S_1|=i$, we can write it as a hockey stick divergence of univariate Gaussian mixture distributions as follows 
\begin{multline}
D_{\alpha}\left(\xi||\xi'\right)\leq 
pqD_{\alpha'}\Big(\sum_{i=0}^{d}\binom{d}{i}q^{i}(1-q)^{d-i}\mathcal{N}((i+1)C,\sigma^{2})\Big|\Big|\bar{c_{1}}\mathcal{N}(0,\sigma^{2})\\+\bar{c_{2}}\sum_{i=0}^{d}\binom{d}{i}q^{i}(1-q)^{d-i}\mathcal{N}(iC,\sigma^{2})\Big),\label{eq:proof_useful_bound}
\end{multline}
where $d$ is the number of elements stored by user 1 excluding $x'$.

Let us define $\mathcal{N}(z,\mu,\sigma^{2})\triangleq\frac{1}{\sigma\sqrt{2\pi}}e^{-\frac{1}{2}(\frac{z-\mu}{\sigma})^{2}}$
and substitute \cref{eq:proof_useful_bound} into \cref{eq:hockey-stick-proof}, which results in
\begin{multline}
D_{\alpha}\left(\xi||\xi'\right) \\\leq
pq\int\Bigg[\sum_{i=0}^{d}\binom{d}{i}q^{i}(1-q)^{d-i}\Big(\mathcal{N}(z,(i+1)C,\sigma^{2}) -\alpha'\bar{c_{2}}\mathcal{N}(z,iC,\sigma^{2})\Big)-\alpha'\bar{c_{1}}\mathcal{N}(z,0,\sigma^{2})\Bigg]_{+}dz.\label{eq:final_bound}
\end{multline}
In \cref{eq:final_bound}, the expression inside $[\cdot]_+$ goes to zero when $z\to \infty$ and $z\to -\infty$. Other than these, the expression has only one zero-crossing $z^*$ and for all finite $z>z^*$, it is positive. Therefore, taking the integral $z\geq z^*$ is enough. Moreover, since there is only one zero-crossing for the expression, it is easy to numerically solve
\begin{equation}
\sum_{i=0}^{d}\binom{d}{i}q^{i}(1-q)^{d-i}\Big(\mathcal{N}(z,(i+1)C,\sigma^{2}) -\alpha'\bar{c_{2}}\mathcal{N}(z,iC,\sigma^{2})\Big)-\alpha'\bar{c_{1}}\mathcal{N}(z,0,\sigma^{2})=0
\end{equation}
for $z$, where $z^*$ is the solution. Therefore, we have
\begin{align}
&D_{\alpha}\left(\xi||\xi'\right) \nonumber\\
&\leq pq\sum_{i=0}^{d}\binom{d}{i}q^{i}(1-q)^{d-i}\Big(1-\Phi_{(i+1)C,\sigma}(z^*) -\alpha'\bar{c_{2}}(1-\Phi_{iC,\sigma}(z^*))\Big)-pq\alpha'\bar{c_{1}}(1-\Phi_{0,\sigma}(z^*))\nonumber \\
&=pq\Bigg(1-\alpha'\bar{c_2}+\alpha'\bar{c_1}(\Phi_{0,\sigma}(z^*)-1)
+\sum_{i=0}^{d}\binom{d}{i}q^{i}(1-q)^{d-i}\Big(\alpha'\bar{c_2}\Phi_{iC,\sigma}(z^*)-\Phi_{(i+1)C,\sigma}(z^*)\Big)\Bigg),\label{eq:final_bound_tidy}
\end{align}
where $\Phi_{\mu,\sigma}$ is the cumulative distribution function of a Gaussian random variable with mean $\mu$ and standard deviation $\sigma$.

Together with \cref{thm:privacy_profile}, \cref{eq:final_bound_tidy} proves
the claim. 

\subsection{Proof of \cref{thm:only_local}}
According to \cref{thm:only-local-sampling}, if a mechanism $\mathcal{M}$ satisfying $(\varepsilon,\delta)$-DP takes as the input a subset sampled from the entire dataset with Poisson sampling, we have $
\delta_{\mathcal{M}'}(\varepsilon')\leq q\delta_{\mathcal{M}}(\varepsilon)
$
and $\varepsilon'=\log\left(1+q(e^{\varepsilon}-1)\right)$, where $\mathcal{M}'$ is the mechanism $\mathcal{M}$ cascaded with sampling. Moreover, from \cref{thm:gaussian_mechanism}, we know $\varepsilon$ vs $\delta$ trade-off for Gaussian mechanism. The claim follows from combining \cref{thm:only-local-sampling} and \cref{thm:gaussian_mechanism}.
\hfill $\square$

\subsection{Proof of \cref{thm:upper_bound}}
The proof can be obtained by following the same steps as in the proof of \cref{thm:main} until \cref{eq:proof_useful_bound}. In \cref{eq:proof_useful_bound} if we ignore $\bar{c_1}\mathcal{N}(0,\sigma^2)$, then we obtain a looser upper bound according to \cref{eq:hockey-stick-proof}. Moreover, ignoring this expression is equivalent to ignoring the contribution of the client sampling when client 1 is sampled. This results in 
\begin{align}
D_{\alpha}\left(\xi||\xi'\right)&\leq pqD_{\alpha'}\Big(\sum_{S_1\in \mathcal{S}_1}\Pr(S_1)\mathcal{N}((i+1)C,\sigma^{2})||
\bar{c_{2}}\sum_{S_1\in \mathcal{S}_1}\Pr(S_1)\mathcal{N}(iC,\sigma^{2})\Big)\\
&\leq \max_{S_1\in\mathcal{S}_1} pqD_{\alpha'}\Big(\mathcal{N}((i+1)C,\sigma^{2})||\bar{c_{2}}\mathcal{N}(iC,\sigma^{2})\Big),\label{eq:proof_useful_bound_loose}
\end{align}
where last inequality is due to the joint convexity of $D_{\alpha'}$. Since it is also shift-invariant, we have
\begin{equation}
D_{\alpha}\left(\xi||\xi'\right)\leq pq\int_{Z}\left[\mathcal{N}(C,\sigma^{2})-\alpha' \bar{c_{2}}\mathcal{N}(0,\sigma^{2})\right]_{+}d(z).\label{eq:loose_bound}
\end{equation}
Since $\varepsilon''=\varepsilon+\log(\bar{c_2})$, the claim follows from \cref{thm:gaussian_mechanism} and \cref{eq:loose_bound}.
\hfill $\square$

\section{Details of the Experimental Setting} \label{sec:experiment_details}

\subsection{Architecture}
\begin{figure}[ht]
\begin{small}
\centering
\resizebox{\linewidth}{!}{
\tikzstyle{relu}=[];
\usetikzlibrary{arrows}
\begin{tikzpicture}

\draw  (-10,2.5) rectangle (-7.5,-0.5);
\draw  (-1.5,2.5) rectangle (1,-0.5);
\draw  (7,2.5) rectangle (9.5,-0.5);
\draw  (12.5,2.5) rectangle (15,-0.5);

\node at (-8.75,3) {Conv. Layer};
\node at (-8.75,2) {Input Ch: 1};
\node at (-8.75,1.5) {Output Ch: 16};
\node at (-8.75,1) {Kernel size: 8};
\node at (-8.75,0.5) {Stride: 2};
\node at (-8.75,0) {Padding: 3};

\node at (-0.25,3) {Conv. Layer};
\node at (-0.25,2) {Input Ch: 16};
\node at (-0.25,1.5) {Output Ch: 32};
\node at (-0.25,1) {Kernel size: 4};
\node at (-0.25,0.5) {Stride: 2};

\node at (8.25,3) {Fully Connected Layer};
\node at (8.25,2) {Input:};
\node at (8.25,1.5) {$32\times 4 \times 4$};
\node at (8.25,0.75) {Output: 256};

\node at (13.75,3) {Fully Connected Layer};
\node at (13.75,2) {Input: 256};
\node at (13.75,1) {Output: 62};

\draw  (-4.5,1.75) rectangle (-2.5,0.25);
\node at (-3.5,2) {MaxPool};
\node at (-3.5,1.25) {Kernel sz: 2};
\node at (-3.5,0.75) {Stride: 1};

\draw  (-6.75,1.5) rectangle (-5.25,0.5);
\node at (-6,1) {ReLU};

\draw  (1.75,1.5) rectangle (3.25,0.5);
\node at (2.5,1) {ReLU};

\draw  (4,1.75) rectangle (6,0.25);
\node at (5,2) {MaxPool};
\node at (5,1.25) {Kernel sz: 2};
\node at (5,0.75) {Stride: 1};

\draw  (10.25,1.5) rectangle (11.75,0.5);
\node at (11,1) {ReLU};

\node at (16.5,1) {Softmax};
\draw  (15.5,1.5) rectangle (17.5,0.5);
\node (v1) at (-7.5,1) {};
\node (v2) at (-6.75,1) {};
\draw [-triangle 60] (v1) edge (v2);
\node (v3) at (-5.25,1) {};
\node (v4) at (-4.5,1) {};
\node (v5) at (-2.5,1) {};
\node (v6) at (-1.5,1) {};
\node (v7) at (1,1) {};
\node (v8) at (1.75,1) {};
\node (v9) at (3.25,1) {};
\node (v10) at (4,1) {};
\node (v11) at (6,1) {};
\node (v12) at (7,1) {};
\node (v13) at (9.5,1) {};
\node (v14) at (10.25,1) {};
\node (v15) at (11.75,1) {};
\node (v16) at (12.5,1) {};
\node (v17) at (15,1) {};
\node (v18) at (15.5,1) {};
\draw [-triangle 60] (v3) edge (v4);
\draw [-triangle 60] (v5) edge (v6);
\draw [-triangle 60] (v7) edge (v8);
\draw [-triangle 60] (v9) edge (v10);
\draw [-triangle 60] (v11) edge (v12);
\draw [-triangle 60] (v13) edge (v14);
\draw [-triangle 60] (v15) edge (v16);
\draw [-triangle 60] (v17) edge (v18);
\end{tikzpicture}}
\caption{Network Architecture Used in the Experiments\label{fig:network}}
\end{small}
\end{figure}
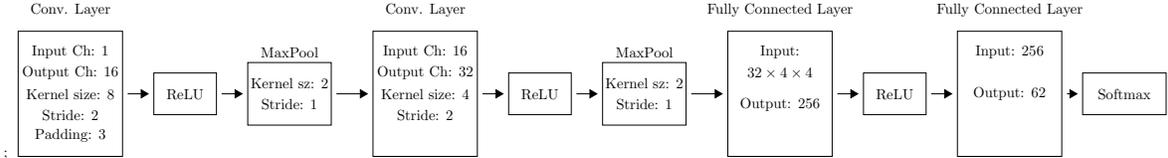

For the experiments in \cref{subsec:small_datasets} and  \cref{subsec:large_datasets} we use a deep neural network with two convolutional layers followed by two fully connected layers. In \cref{fig:network}, we give the visualization of the network with all necessary details.

\subsection{Hyperparameters}
In \cref{subsec:small_datasets}, we use SGD optimizer with a momentum 0.9. Learning rate for the non-private case and the proposed method is 0.002. For \UpperBoundAcr and \OnlyLocalAcr, we observed the learning rate should be smaller due to the high noise addition. Hence, the learning rate of 0.0001 is used for these schemes. 

In \cref{subsec:large_datasets}, we used SGD optimizer with a momentum of 0.9. The learning rate is 0.002, for all schemes.

In all experiments, we used an average batch size of 70.

\end{document}